\documentclass[conference]{IEEEtran}
\IEEEoverridecommandlockouts
\usepackage{cite}
\usepackage{amsmath,amssymb,amsfonts}
\usepackage{graphicx}
\usepackage{booktabs}
\usepackage{array}
\usepackage{textcomp}
\usepackage{xcolor}
\usepackage{url}
\usepackage{balance}
\usepackage[hidelinks]{hyperref}
\graphicspath{{figs/}}

\newcommand{\Mone}{\mathcal{M}^{1}}
\newcommand{\Mtwo}{\mathcal{M}^{2}}
\newcolumntype{L}[1]{>{\raggedright\arraybackslash}p{#1}}
\newlength{\singlegap}
\newlength{\singlew}

\begin{document}

\title{Shared-Policy Dual-Humanoid Object Carrying via an End-to-End Whole-Body Control Pipeline}

\author{\IEEEauthorblockN{Jie Wu}
\IEEEauthorblockA{Department of Mechanical Engineering and Applied Mechanics\\
University of Pennsylvania\\
Philadelphia, PA, USA\\
wu3je@engineering.upenn.edu}
}

\maketitle

\begin{abstract}
Humanoids that manipulate objects in human environments must coordinate locomotion, posture, arm motion and contact; when two humanoids carry one object, their motions are additionally coupled through the object. We present an end-to-end pipeline that turns human--object and human--object--human motion capture into whole-body tracking policies for the 29-DoF Unitree G1, trained with PPO in Isaac Sim: source motions are converted to a common convention, retargeted with an interaction-preserving retargeter, turned into object assets and references, and tracked by policies in which the object is simulated dynamically. We validate the single-humanoid path on eight OMOMO objects and extend it to two-person carrying from CORE4D with one \emph{shared} 29-DoF policy: the actor is evaluated once per robot, both actions are executed in the same scene, each robot contributes its own on-policy sample, and coordination emerges from the synchronized references and the physical object. A synchronized dual-person retargeting procedure keeps a common floor, scale and object trajectory. Three object classes with different contact geometry---box, bucket and chair---are learned with one common tracking reward plus a few demonstration-gated, class-conditional terms (hand--object proximity, a wider contact gate, a feet--footprint penalty) and scene-level termination rules. We report training curves, tracking statistics and a diagnosed case study of the chair, including the alternatives that were discarded.
\end{abstract}


\section{Introduction}
Humanoid robots can potentially operate in environments designed for people, but everyday object interaction requires more than stable locomotion: the robot must coordinate locomotion, posture, arm motion and contact while controlling the pose of an external object. The task becomes substantially harder when two humanoids manipulate the same object: their motions are coupled through contact---an error by either robot changes the object's motion and thereby the dynamics observed by both---so a collaborative controller must solve whole-body tracking and physical cooperation at once.

\begin{figure}[!t]
\centering
\includegraphics[width=0.94\columnwidth]{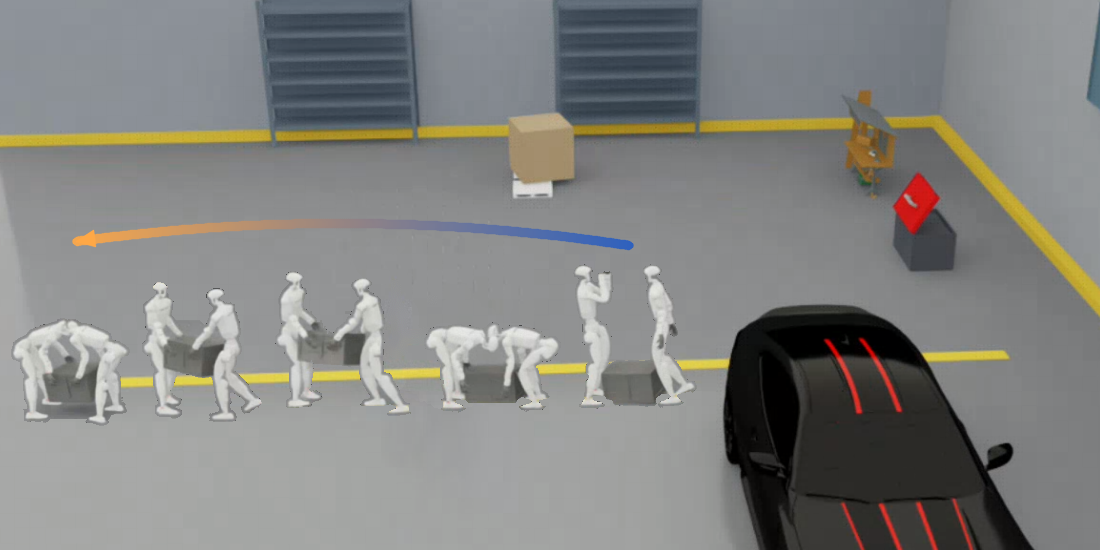}
\caption{Two Unitree G1 humanoids driven by one shared 29-DoF policy carry a box (CORE4D-Real \texttt{20231011/034}, \texttt{box021}), the reference class of the common recipe. Robots and box are cut out of five key frames of one Isaac Sim evaluation rollout---approach, pick-up, carry, lowering, release---and placed on a clean plate of the scene in temporal order under the object-trajectory arrow (blue: start, orange: end); frame spacing is chosen for legibility and does not preserve the true displacement. The garage is a visual-only background shared by all garage figures.}
\label{fig:teaser}
\end{figure}

Motion-imitation methods have produced increasingly capable humanoid controllers, and human--object interaction datasets provide rich references for carrying and rearrangement. Converting captured two-person interaction data into a working robot controller, however, remains a multi-stage problem: motions must be transformed into a common convention, retargeted to a robot with different proportions and joint limits, synchronized with the object, and converted into a representation that a reinforcement-learning environment can consume---and for collaborative tasks the two independently retargeted motions must stay aligned with one shared object in one physical scene.

This paper presents an end-to-end pipeline from human--object motion data to whole-body control policies for the 29-DoF Unitree G1, trained with PPO~\cite{schulman2017ppo}. We first establish the pipeline in the single-humanoid setting on eight OMOMO objects---a large box, a trash can, a suitcase, a tripod, two chairs, a floor lamp and a small table---with one reward, termination and simulation setup, and then extend it to two-humanoid carrying with two-person motions from CORE4D. Rather than learning two specialized policies or a monolithic 58-dimensional controller, we use \emph{one shared} 29-DoF policy for both robots: in each scene the policy is evaluated once per humanoid, and the two actions drive two independently simulated G1 robots that interact with the same dynamic object. This preserves the single-robot action space, shares experience between the roles, and lets coordination emerge through the common reference and the physical coupling induced by the object. The extension is demonstrated on three CORE4D object classes with different contact geometry: a box, a bucket and a chair (Figs.~\ref{fig:teaser}, \ref{fig:dual} and \ref{fig:chair}).

The contribution is not a new learning algorithm but a reproducible motion-to-control pipeline and its extension to a shared-policy, physically coupled two-humanoid setting: (i)~an end-to-end pipeline from motion conversion and retargeting to asset generation, PPO training and evaluation; (ii)~a shared-policy formulation in which two G1 robots are controlled by a single actor and interact with one shared dynamic object; (iii)~a synchronized dual-person retargeting procedure preserving a common frame, floor, scale, time base and object trajectory; (iv)~class-conditional reward and termination terms for collaborative carrying, with the diagnosed failures that motivated them; and (v)~validation on eight single- and three dual-humanoid object--motion pairs.

\section{Related Work}
\textbf{Physics-based motion imitation.} DeepMimic showed that reinforcement learning can reproduce motion capture with pose, velocity and task objectives~\cite{peng2018deepmimic}; adversarial motion priors and large motion libraries improved robustness and diversity~\cite{peng2021amp,luo2023phc}, and recent whole-body trackers transfer dynamic skills to robot morphologies~\cite{liao2025beyondmimic}. Our reference additionally includes a dynamic object whose physical evolution must be reproduced, and a second humanoid acting on the same object.

\textbf{Human--object interaction data and control.} OMOMO provides full-body interactions with household objects~\cite{li2023object}; InterMimic processes such motions into a common representation and learns physics-based control over many interaction skills~\cite{xu2025intermimic}. These supply the source motions and object assets of our single-humanoid stage; motions are moved onto the robot with the interaction-preserving retargeter OmniRetarget~\cite{yang2025omniretarget}. CORE4D provides synchronized human--object--human sequences for collaborative rearrangement, including two-person carrying~\cite{liu2025core4d}.

\textbf{Multi-agent and collaborative control.} CooHOI learns an individual carrying skill with an adversarial motion prior and transfers it to cooperative transport with multi-agent PPO, using object dynamics as an implicit communication channel~\cite{gao2024coohoi}; TeamHOI learns a decentralized shared policy for variable team sizes with a Transformer and explicit teammate tokens~\cite{lionar2026teamhoi}. Concurrent human--humanoid and human--robot transport work studies multi-agent cognition-to-control, safety-aligned gradients, interaction-aware compliant carrying and heterogeneous-agent Lyapunov optimization~\cite{zhang2026cognition,li2026sage,zhang2026interaction,zhang2026halo}. We share parameters between two identical humanoids: each robot produces one 29-DoF action through the same MLP actor from an observation in its own role. Unlike CooHOI we track synchronized two-person demonstrations; unlike TeamHOI we specialize to two robots and a compact shared MLP.

\section{Motion-to-Control Pipeline}
\label{sec:pipeline}
The pipeline maps a recorded interaction to a G1 policy in seven steps: (1)~acquire synchronized human and object trajectories; (2)~normalize body format, coordinate convention, scale, floor height and frame rate; (3)~retarget to the 29-DoF G1; (4)~convert to the 50\,Hz whole-body-tracking (WBT) representation; (5)~build a dynamic object asset; (6)~train with PPO in Isaac Sim through Isaac Lab~\cite{mittal2023orbit}; (7)~evaluate with robot and object simulated dynamically. Single- and dual-humanoid paths share steps 3--7.

\subsection{Single-Humanoid Data}
\begin{figure*}[!t]
\centering
\setlength{\singlegap}{0.016\textwidth}
\setlength{\singlew}{\dimexpr0.25\textwidth-0.75\singlegap\relax}
\begin{tabular}{@{}c@{\hspace{\singlegap}}c@{\hspace{\singlegap}}c@{\hspace{\singlegap}}c@{}}
\includegraphics[width=\singlew]{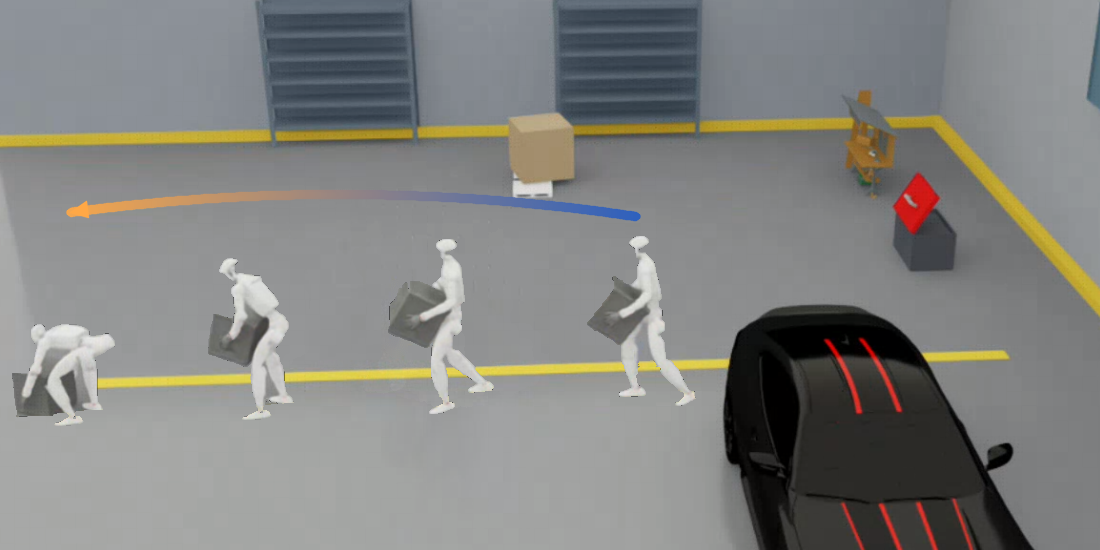}&
\includegraphics[width=\singlew]{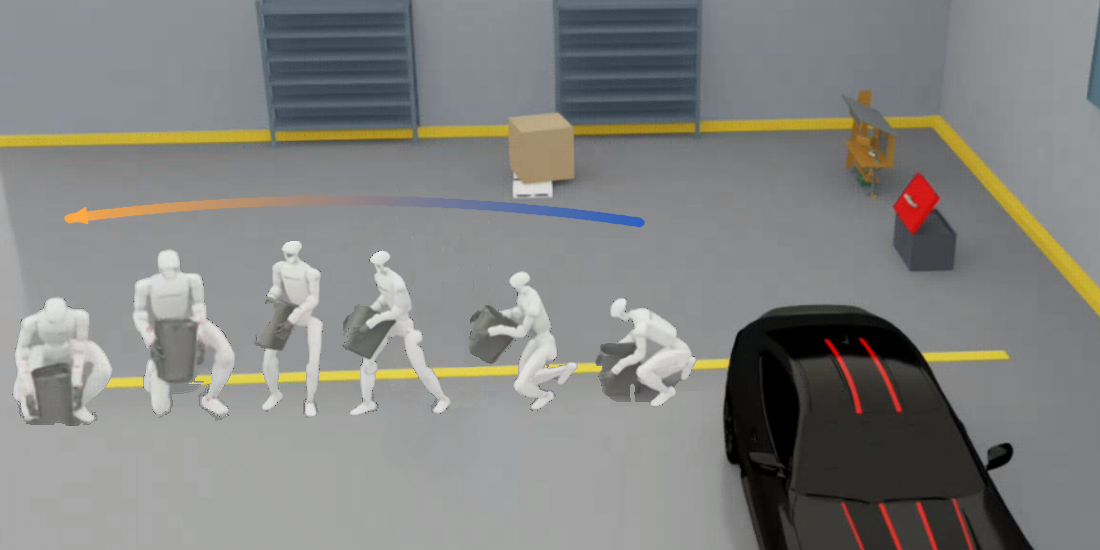}&
\includegraphics[width=\singlew]{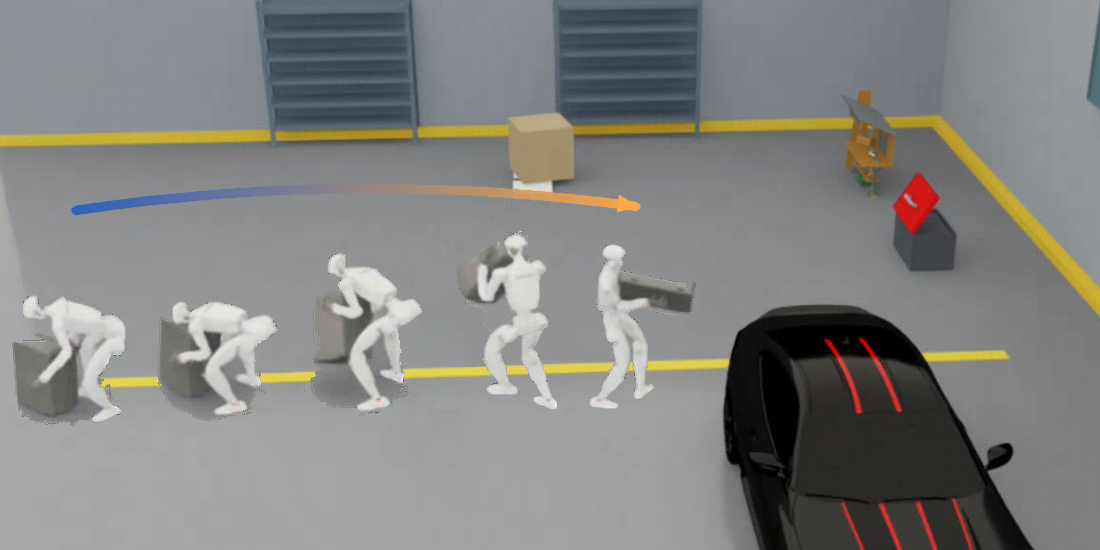}&
\includegraphics[width=\singlew]{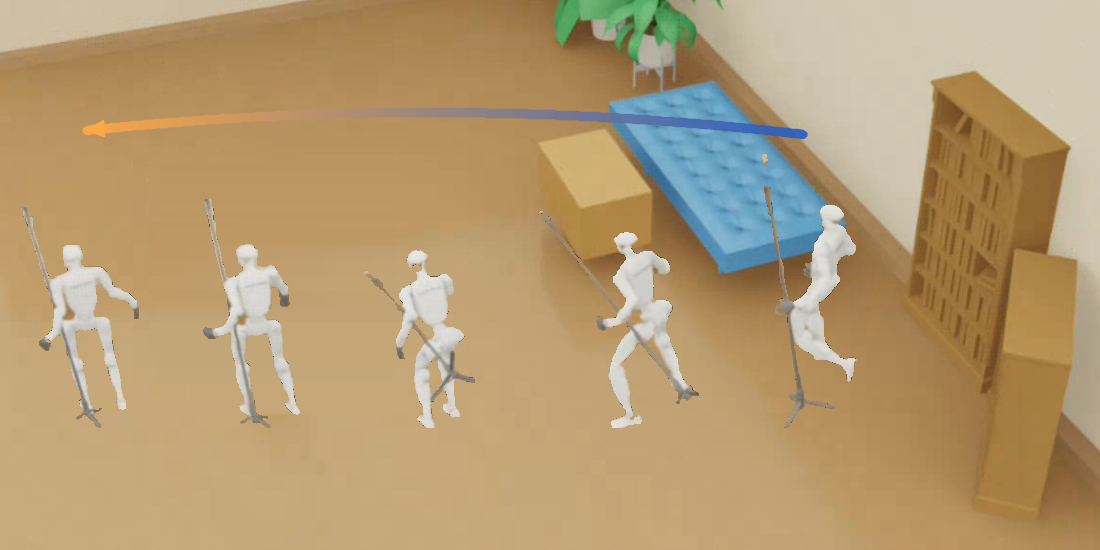}\\[\singlegap]
\includegraphics[width=\singlew]{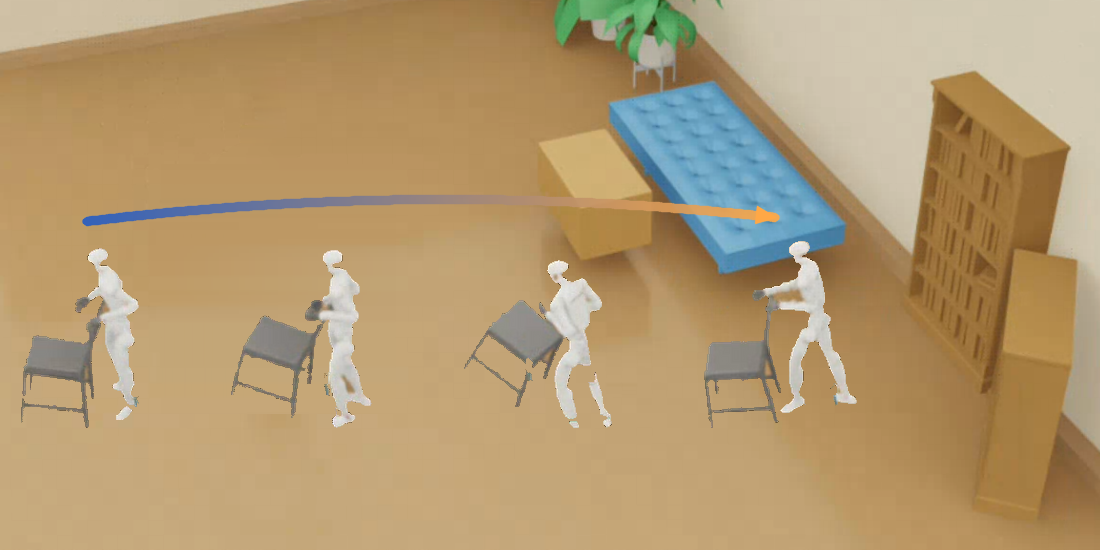}&
\includegraphics[width=\singlew]{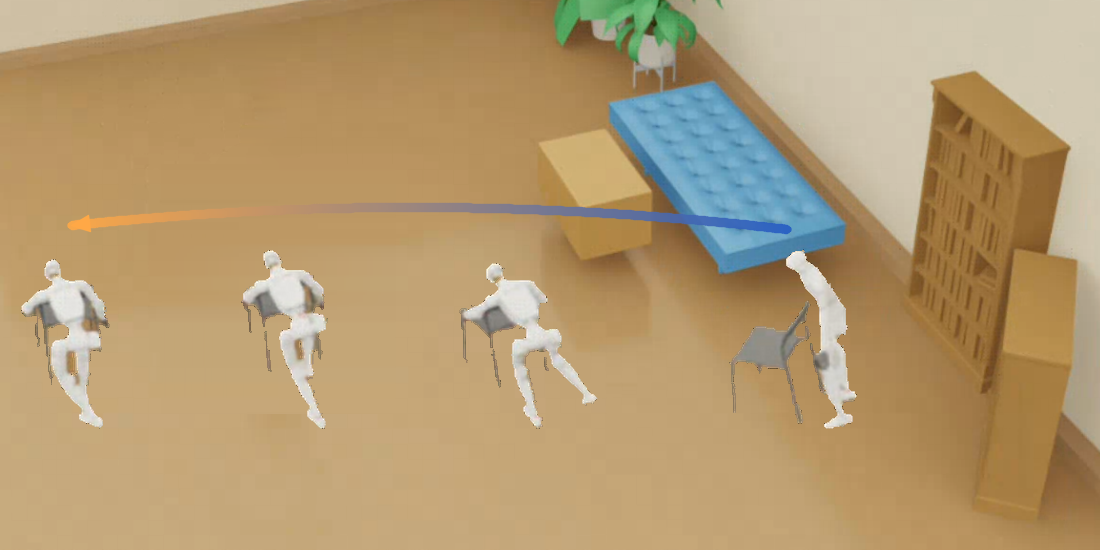}&
\includegraphics[width=\singlew]{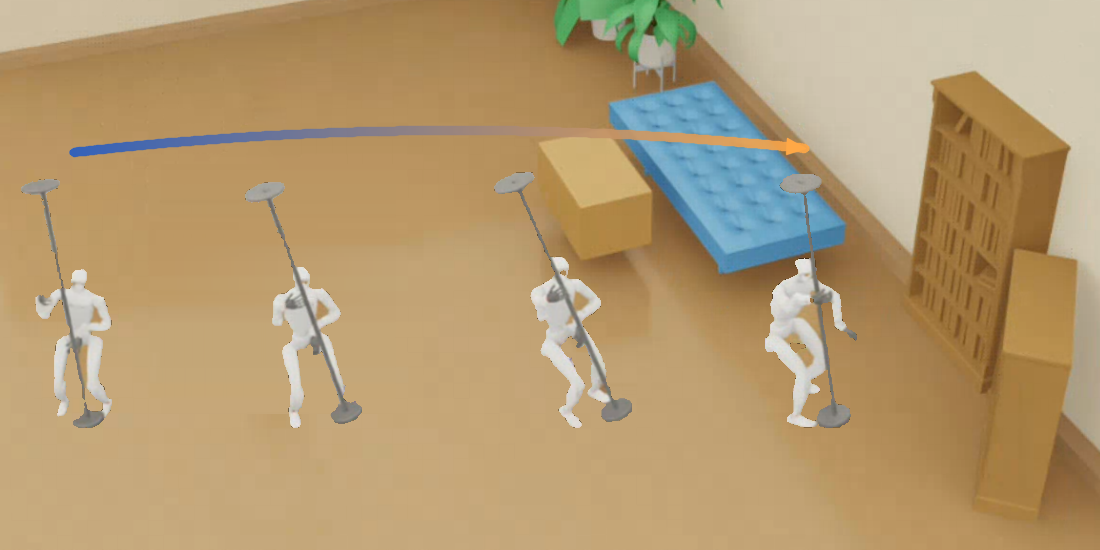}&
\includegraphics[width=\singlew]{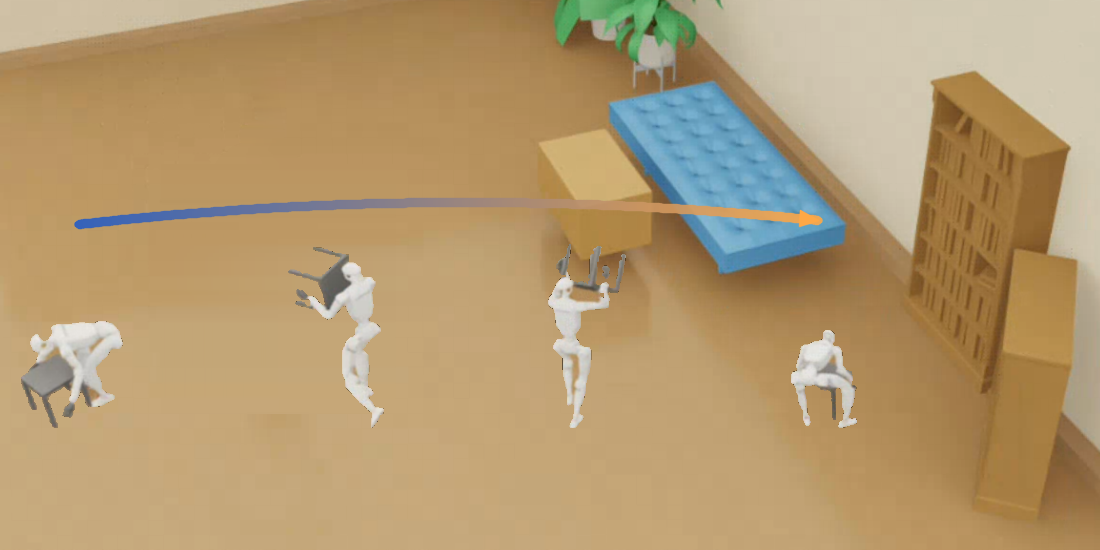}
\end{tabular}
\caption{The eight OMOMO objects of the single-humanoid stage: large box, trash can, suitcase, tripod (top); white chair, wooden chair, floor lamp, small table (bottom). Each panel is one evaluation rollout of the corresponding PPO policy, with four to six key frames placed on the shared plate of its scene at a common scale; the arrow marks the object trajectory from start (blue) to end (orange). Garage and living room are visual-only backgrounds.}
\label{fig:single}
\end{figure*}

We use OMOMO motions in the InterMimic representation (a 591-dimensional record with the SMPL-H body and the object trajectory), one clip per object (Table~\ref{tab:single}). Each motion is retargeted with the object-interaction mode of OmniRetarget, which resolves the differences in proportions, topology and joint ranges while preserving the body--object relation, and resampled to the 50\,Hz WBT format (root and joint trajectories, tracked body poses and velocities, synchronized object pose and velocity). One policy is trained from scratch per clip. Fig.~\ref{fig:single} shows the eight objects and one evaluation rollout of each policy.

\subsection{Dual-Person CORE4D Conversion}
CORE4D-Real provides two synchronized SMPL-X subjects, one object trajectory and the object geometry, in a Y-up frame at 15\,Hz. We map the SMPL-X joints to the 52-joint SMPL-H layout expected by the retargeter, transform points from $(x,y,z)$ Y-up to $(x,-z,y)$ Z-up together with the object rotations, and resample bodies and object to 30\,Hz (linear interpolation for positions, slerp for orientations). A \emph{single} floor offset is computed jointly from both subjects and the object and subtracted from all three trajectories, and one shared height estimate is used for both participants; estimating these separately would shift the two retargeted robots into incompatible coordinate systems. Each person is then packed into a separate InterMimic-style record with the identical object trajectory, retargeted independently with the same object model and scale, and converted to 50\,Hz, giving motion files $\Mone=\{q^1_t,x^1_t,x^o_t\}$ and $\Mtwo=\{q^2_t,x^2_t,x^o_t\}$ with joint configurations $q^i_t$, tracked body states $x^i_t$ and the shared object trajectory $x^o_t$. We verify that both robots stand in one world frame on opposite sides of the object (Fig.~\ref{fig:refs}, top).

\subsection{Object Assets and Classes}
Each mesh becomes a simulator asset with visual and collision geometry, a free root joint and a nominal mass of 0.1\,kg, used unchanged by conversion, training and evaluation. In dual-humanoid scenes one dynamic object is instantiated and both robots collide with it; there is no weld, magnetic grasp or hand--object joint, so the object is the physical communication channel between the robots. Its PhysX friction is set to 1.6/1.2 (static/dynamic) before randomization, because the generic object range (0.1--0.6) is far too slippery for a 0.1\,kg object held between two pairs of hands.

\begin{table}[!t]
\caption{CORE4D-Real two-person clips (50\,Hz). Reset $z$: lowest object-centre height at reset.}
\label{tab:clips}
\centering\footnotesize
\setlength{\tabcolsep}{3.5pt}
\begin{tabular}{@{}llccc@{}}
\toprule
Class & Object & Size $l\times w\times h$ (m) & Frames / s & Reset $z$ (m)\\
\midrule
Box & \texttt{box021} & $0.39\times0.30\times0.50$ & 750 / 15.0 & 0.26\\
Bucket & \texttt{bucket007} & $0.55\times0.57\times0.57$ & 380 / 7.6 & 0.29\\
Chair & \texttt{chair006} & $0.59\times0.56\times0.86$ & 577 / 11.5 & 0.43\\
\bottomrule
\end{tabular}
\end{table}

From the CORE4D carrying sequences (\texttt{20231011/034}, \texttt{20231018/019}, \texttt{20231003\_2/011}) we group objects by contact geometry and train one clip per class (Table~\ref{tab:clips}). The \emph{box} is compact and both people place their palms on opposing faces near the volume centre; it is the reference case for the common recipe. The \emph{bucket} is a lidded container held at the rim, so the demonstrated wrists are offset from the centroid and the grasp can slip along the rim. The \emph{chair} has a large floor footprint; its grasp points on the backrest rail and the front of the frame lie 0.46--0.50\,m from the mesh centre (Fig.~\ref{fig:refs}, bottom), and the demonstration lifts it only about 0.19\,m.

\section{Shared-Policy Dual-Humanoid Control}
\label{sec:control}
\subsection{Formulation and Shared Sampling}
\begin{figure}[!t]
\centering
\includegraphics[width=0.94\columnwidth]{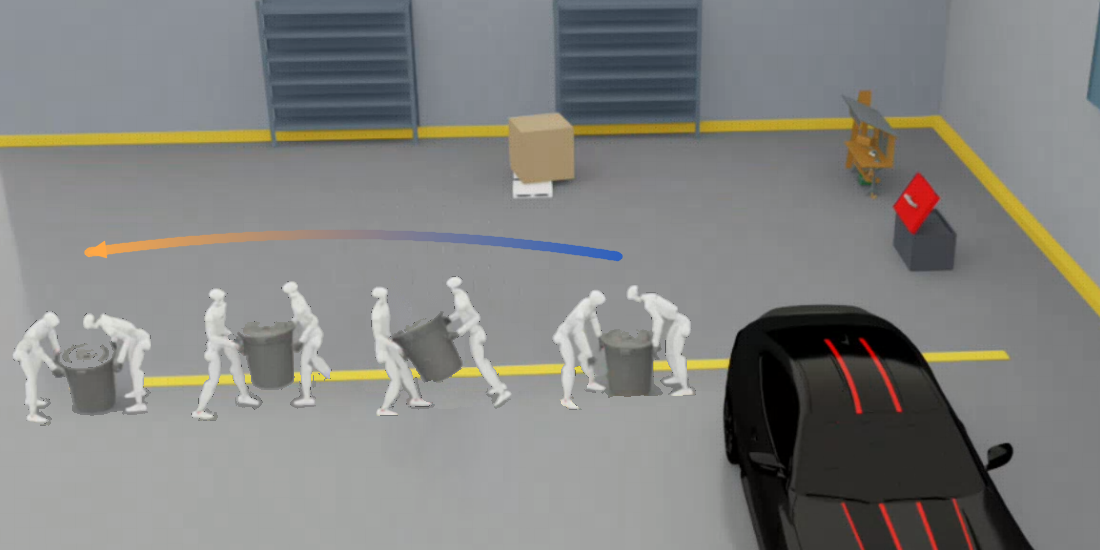}
\caption{Shared-policy carrying of the bucket (\texttt{bucket007}). Four key frames of one evaluation rollout; the bucket is held at the rim and the demonstration-gated hand--object proximity term keeps the wrists at the demonstrated horizontal distance to its centre. Arrow: object trajectory from start (blue) to end (orange).}
\label{fig:dual}
\end{figure}

Two identical 29-DoF G1 robots $i\in\{1,2\}$ and one dynamic object share a scene. At step $t$ robot $i$ receives a role-conditioned observation $o^i_t$ and draws $a^i_t\sim\pi_\theta(a\mid o^i_t)$ with common parameters $\theta$; the simulator executes $(a^1_t,a^2_t)$ jointly, $s_{t+1}=f(s_t,a^1_t,a^2_t;\Mone,\Mtwo)$. The objective is to track both retargeted body motions while reproducing the shared object trajectory without falling or losing the interaction. With $N$ scenes a wrapper exposes $2N$ agent samples to PPO: the observations of both persons are concatenated along the batch dimension, passed through the same actor and critic, and split back into two action batches; rewards are stacked in the same order, while termination is shared within a scene because failure of either participant or of the object invalidates the collaborative rollout. We use $N=1{,}024$ scenes, i.e.\ 2{,}048 PPO samples per step as in the single-humanoid setup. This is parameter-shared independent-agent PPO, not centralized training with decentralized execution: neither actor nor critic sees the other humanoid's state, while the two samples remain physically coupled through the object. Fig.~\ref{fig:dual} shows the shared policy on the bucket; the box is Fig.~\ref{fig:teaser} and the chair is Fig.~\ref{fig:chair}.

\subsection{Observations, Actions and PPO}
The actor observation (154 dimensions) contains the reference-conditioned motion command, the reference orientation in the robot frame, base angular velocity, joint positions and velocities, and the previous action, all computed from the robot's own reference. The critic (298 dimensions) additionally receives privileged information: reference position, tracked body poses, base linear velocity, and the shared object's pose and linear velocity relative to the current robot. The actor receives neither a partner state nor an object state; coordination is induced by the synchronized references, the shared parameters and the object's physical response felt through contact and proprioception. Actions are 29 joint-position targets converted to torques by a PD controller; physics runs at 200\,Hz with decimation four, i.e.\ a 50\,Hz policy matching the references.

The shared actor and critic are optimized with the standard clipped PPO objective~\cite{schulman2017ppo}, with the expectation taken over both humanoids and time and the generalized advantage of each sample computed from that robot's own reward and value. All PPO policies use the same settings: 512--256--128 ELU MLPs with observation normalization; 24-step rollouts from 2{,}048 agents (49{,}152 samples per iteration), 5 epochs over 4 minibatches; $\gamma=0.99$, $\lambda=0.95$, clip 0.2, entropy coefficient 0.005, gradient-norm clip 1.0; a KL-adaptive learning rate (initial $10^{-3}$, target KL 0.01) and initial action standard deviation 1.0. A dual-humanoid iteration takes 3.1\,s on a 12\,GB laptop GPU ($\approx$8.6\,h per 10k iterations).

\subsection{Reset, Synchronization and Termination}
\label{sec:term}
At reset both robots are initialized from the same time index of their clips and the object from the shared trajectory; the start index is drawn by an adaptive sampler that favours segments in which episodes recently failed. The references advance with a common clock; when a clip ends the scene is re-initialized from a new start index without ending the PPO episode, which ends only on a failure or at the 10\,s timeout (500 steps). The carry configuration uses reduced initial perturbations (joints $\pm0.08$\,rad, root position $\pm0.03$\,m and object position $\pm0.02$\,m in $xy$, root orientation $\pm(0.08,0.08,0.12)$\,rad, no vertical velocity), zeroes the object velocity and clamps the object centre to the reset height of Table~\ref{tab:clips}; with the generic perturbation the 0.1\,kg box was kicked away when both robots landed at spawn.

Termination is evaluated per scene and applied to both robots: (a)~torso height error $>0.5$\,m or horizontal error $>0.25$\,m; (b)~projected-gravity tilt error $>0.8$; (c)~height error of any tracked ankle or wrist $>0.25$\,m; (d)~object centre off its reference by $>0.25$\,m or object orientation error $>0.8$\,rad; (e)~a \emph{contacting} wrist misses the object: whenever the demonstration's closer wrist is within a gate $d_{\mathrm{gate}}$ of the object centre, the simulated wrist may exceed the demonstrated horizontal wrist--object distance by at most 0.20\,m. The horizontal part of (a) and rule (e) are specific to carrying: the single-humanoid tracker checks heights only, so a robot standing beside the box at the correct height, or a hand hovering next to the object, would never be reset---Fig.~\ref{fig:curves} shows this degenerate solution for the box before the rules were enabled. $d_{\mathrm{gate}}$ is 0.45\,m for box and bucket and 0.60\,m for the chair (Section~\ref{sec:chair}).

\subsection{Reward Design}
\label{sec:reward}
All policies share the body-and-object tracking reward of Table~\ref{tab:reward}. Body terms compare each robot with its own reference; object terms compare the simulated object with the shared trajectory and are identical for both robots of a scene. Tracking terms have the form $\exp(-e^2/\sigma^2)$ and regularizers are quadratic penalties. With the termination rules this common reward suffices for the box. Two additional terms were introduced for the bucket and the chair; both are \emph{demonstration-gated}, i.e.\ active only while the demonstration itself is in contact with the object, so approach and retreat are untouched.

\begin{table}[!t]
\caption{Common reward terms.}
\label{tab:reward}
\centering\footnotesize
\setlength{\tabcolsep}{3.5pt}
\begin{tabular}{@{}L{2.8cm}L{2.35cm}cc@{}}
\toprule
Term & Error $e$ & $\sigma$ & Weight\\
\midrule
Global ref.\ position / orientation & torso pose vs.\ reference & 0.3\,m / 0.4\,rad & 0.5\\
Relative body positions / orientations & 14 bodies, torso frame & 0.3\,m / 0.4\,rad & 1.0\\
Body linear / angular velocities & 14 bodies & 1.0\,m/s / 3.14\,rad/s & 1.0\\
Object position / orientation & object pose vs.\ reference & 0.3\,m / 0.4\,rad & 1.0\\
Action rate & $\lVert a_t-a_{t-1}\rVert^2$ & -- & $-0.1$\\
Soft joint limits & beyond 90\,\% of range & -- & $-10$\\
Undesired contacts & force $>1$\,N off feet, ankles, wrists & -- & $-0.1$\\
\bottomrule
\end{tabular}
\end{table}

\begin{figure*}[!t]
\centering
\includegraphics[width=0.96\textwidth]{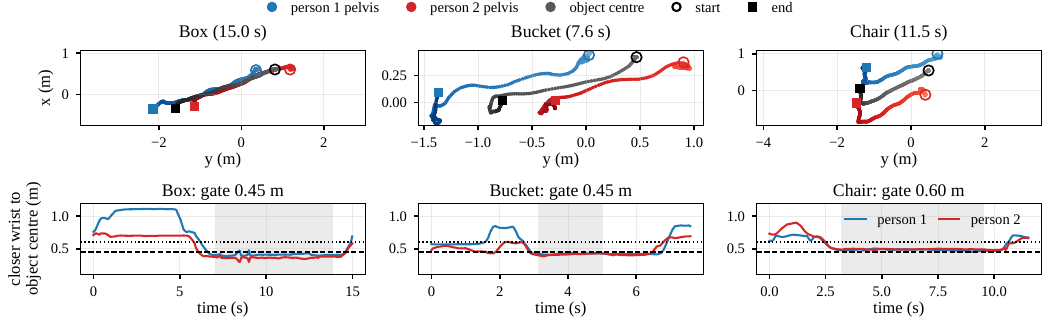}
\caption{Converted two-person references (50\,Hz). Top: top-down trajectories of the pelvis of person~1 (blue), person~2 (red) and the object centre (grey), light to dark in time; circles mark the start, squares the end. Bottom: demonstrated distance from the object centre to the closer wrist of each person, with the carry phase (object lifted $>3$\,cm) shaded and the two contact gates drawn (dashed 0.45\,m, dotted 0.60\,m). Box and bucket wrists sit 0.34--0.43\,m from the centre, inside the 0.45\,m gate; the chair is held 0.46--0.50\,m away for the entire carry, so the soft gate $g_t$ of (\ref{eq:hand}) is only 0.27--0.38 and the 0.45\,m termination gate never arms.}
\label{fig:refs}
\end{figure*}

\textbf{Demonstration-gated hand--object proximity (bucket, chair).} Let $d^{\mathrm{demo}}_{3D,t}$ be the distance from the demonstrated object centre to the closer demonstrated wrist, and $d^{\mathrm{demo}}_{xy,t}$, $d^{\mathrm{sim}}_{xy,t}$ the corresponding horizontal distances in demonstration and simulation. The term rewards a horizontal grasp at least as close as the demonstration,
\begin{equation}
r^{\mathrm{hand}}_t=g_t\exp\!\Big(-\tfrac{\max(0,\,d^{\mathrm{sim}}_{xy,t}-d^{\mathrm{demo}}_{xy,t})^2}{\sigma_h^2}\Big),\;
g_t=\sigma\!\Big(\tfrac{d_{\mathrm{gate}}-d^{\mathrm{demo}}_{3D,t}}{\kappa}\Big)
\label{eq:hand}
\end{equation}
where $\sigma_h=0.08$\,m, $\kappa=0.04$\,m, weight 1.0 and the class gate $d_{\mathrm{gate}}$; $\sigma(\cdot)$ is the logistic function. Only horizontal slack is penalized: a hand above the rim or the seat is not a grasp. Fig.~\ref{fig:refs} (bottom) plots the demonstrated wrist--object distances against both gates and shows why the chair is outside the 0.45\,m box/bucket gate for the entire carry.

\textbf{Gated feet--footprint penalty (chair).} Each ankle position is expressed in the chair's mesh frame and compared with the floor footprint, an axis-aligned box with half-extents $(0.297,0.28)$\,m. With $d_{\mathrm{out}}$ the signed distance to the nearest footprint edge (negative inside), each foot contributes $c_t=\mathrm{clamp}((m-d_{\mathrm{out}})/m,0,1)$ with $m=0.10$\,m, i.e.\ 1 inside the footprint fading to 0 at 0.10\,m outside. The sum over both feet is weighted by $-1.0$ and by two gates: the first is 1 only after the demonstration has lifted the chair more than 0.03\,m (the approach is not pushed backwards); the second is 1 while the simulated chair is within 0.10\,m of its rest height and fades to 0 by 0.20\,m above it, so only feet under a chair about to touch down are penalized and the policy may also escape the penalty by lifting higher. The demonstrated persons stand 0.11--0.19\,m clear of the seat edge, so the demonstration itself is essentially never penalized.

\subsection{Domain Randomization}
At environment creation the robot's static/dynamic friction is sampled in $[0.3,1.6]$/$[0.3,1.2]$ and restitution in $[0,0.5]$, the base centre of mass is offset by up to $\pm0.025$\,m ($x$) and $\pm0.05$\,m ($y,z$), and each joint receives a bias in $\pm0.01$\,rad; the base is pushed every 1--3\,s with velocity changes up to $(0.5,0.5,0.2)$\,m/s and $(0.52,0.52,0.78)$\,rad/s. The object mass is scaled by $[1,4]$ and $I_{xx}$ by $[0.5,1.5]$. Single-humanoid runs sample the object friction in $[0.1,0.6]$ as in the reference tracker; the collaborative ``grip'' preset samples static/dynamic friction in $[0.9,1.6]$/$[0.8,1.4]$ and restitution in $[0,0.15]$.

\section{Experiments}
\subsection{Setup and Protocol}
All policies were trained on one laptop GPU (NVIDIA RTX~4080 Laptop, 12\,GB); single-humanoid policies use the half-sphere hand collision model (rubber-hand for the small table), dual-humanoid policies the rubber-hand model. Evaluation runs a checkpoint in a single Isaac Sim GUI scene with the same motion files and object URDF and a visual-only garage or living-room background; both robots of a dual scene use the same checkpoint. A policy counts as a success when the object is picked up, carried along the demonstrated path and set down without a fall, drop or reset in repeated rollouts; the composites of Figs.~\ref{fig:teaser}, \ref{fig:single}, \ref{fig:dual} and \ref{fig:chair} cut robot and object out of four to six screenshots of such a rollout and place them at one common scale on a clean plate of the scene. \emph{Training-time statistics} are means over the last 200 logged iterations under the full training distribution (adaptive start indices, randomization, pushes): mean episode length in control steps (timeout 500; clip ends do not end an episode, failures do), torso position error, mean position error of the 14 tracked bodies relative to the torso, and joint-position error. \emph{Diagnostics} for the chair roll out 256 training-configured scenes from the clip start and record when and why each scene terminated; they keep the training randomization, are therefore pessimistic (about 30\,\% of box scenes terminate within the first second), and only compare stages of one object.

\subsection{Single-Humanoid Validation}
\label{sec:single}
\begin{table}[!t]
\caption{Single-humanoid PPO policies (mean of last 200 iterations). Clip: OMOMO $N$/$k$; torso/body in m, joints in rad.}
\label{tab:single}
\centering\footnotesize
\setlength{\tabcolsep}{3pt}
\begin{tabular}{@{}llcccccc@{}}
\toprule
Object & Clip & Frames / s & Iters & Ep. & Torso & Body & Joints\\
\midrule
Large box & 3/003 & 325 / 6.5 & 30k & 334 & 0.120 & 0.059 & 1.14\\
Trash can & 6/006 & 390 / 7.8 & 34k & 344 & 0.140 & 0.064 & 1.11\\
Suitcase & 17/006 & 377 / 7.5 & 30k & 359 & 0.129 & 0.061 & 1.10\\
Floor lamp & 10/007 & 390 / 7.8 & 30k & 308 & 0.147 & 0.069 & 1.37\\
Small table & 17/019 & 249 / 5.0 & 30k & 214 & 0.135 & 0.063 & 1.22\\
\midrule
Tripod & 10/042 & 242 / 4.8 & 30k & 291 & 0.143 & 0.067 & 1.33\\
White chair & 10/000 & 179 / 3.6 & 30k & 268 & 0.131 & 0.062 & 1.19\\
Wooden chair & 12/000 & 397 / 7.9 & 30k & 348 & 0.126 & 0.060 & 1.15\\
\bottomrule
\end{tabular}
\end{table}

Table~\ref{tab:single} lists the eight object--motion pairs (clip durations 3.6--7.9\,s); the corresponding rollouts are Fig.~\ref{fig:single}. All eight use the reward of Table~\ref{tab:reward}, the height-only termination of the reference tracker and the same assets and simulation setup, and were trained with PPO for 30{,}000 iterations (34{,}000 for the trash can). Training-time torso error is 0.12--0.15\,m and the relative body error 0.06--0.07\,m (Fig.~\ref{fig:errors}); the small table has the shortest mean episode length (214 steps). All eight complete their clip in GUI evaluation; a small-box clip trained with the same recipe did not and is excluded.

\subsection{Dual-Humanoid Carrying}
\label{sec:dual}
\begin{table}[!t]
\caption{Collaborative training stages (each resumes from the previous checkpoint). Base: height-only termination; carry: adds the torso-XY and wrist rules of Section~\ref{sec:term}.}
\label{tab:stages}
\centering\footnotesize
\setlength{\tabcolsep}{3.5pt}
\begin{tabular}{@{}llL{3.35cm}L{2.15cm}@{}}
\toprule
Class & Iterations & Reward & Termination\\
\midrule
Box & 0--8k & common + side-face clamp (w\,2.0, $\sigma$\,0.05\,m) & base\\
Box & 8k--30k & common & base\\
Box & 30k--50k & common & carry, 0.45 (final)\\
\midrule
Bucket & 0--28k & common & carry, 0.45\\
Bucket & 28k--40k & common + hand proximity (0.45) & carry, 0.45 (final)\\
\midrule
Chair & 0--20k & common + hand proximity (0.45) & carry, 0.45\\
Chair & 20k--30k & common + hand proximity (0.60) & carry, 0.60\\
Chair & 30k--40k & + object-height term & carry, 0.60\\
Chair & 40k--50k & + feet--footprint, no height term & carry, 0.60 (final)\\
\bottomrule
\end{tabular}
\end{table}

\begin{figure*}[!t]
\centering
\includegraphics[width=0.96\textwidth]{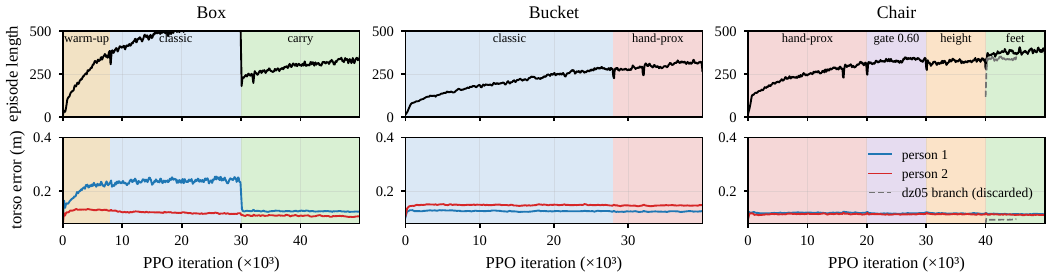}
\caption{Training curves of the three collaborative policies (200-iteration moving average) with the stages of Table~\ref{tab:stages} shaded. Top: mean episode length (timeout 500). Bottom: torso position error of the two robots. Under the height-only ``base'' termination the box episodes run to the timeout while person~1 stands 0.25\,m off its reference; enabling the carry rules at 30k shortens the episodes and halves that error. The discarded chair branch is dashed.}
\label{fig:curves}
\end{figure*}

Each collaborative policy was trained in stages, resuming from the previous checkpoint whenever a reward or termination term changed (Table~\ref{tab:stages}, Fig.~\ref{fig:curves}); all stages use 1{,}024 scenes, the carry reset configuration, the grip randomization and the object friction of Section~\ref{sec:pipeline}.

\textbf{Box.} The first two stages used the single-humanoid termination, and Fig.~\ref{fig:curves} (left) shows the consequence: the mean episode length saturates at the 500-step timeout while the torso error of person~1 stays at 0.25\,m---the policy kept one robot beside the box at the correct height, which height-only checks cannot reset. Enabling the carry rules at 30k immediately terminates these episodes, and person~1's torso error drops to 0.12\,m within a few hundred iterations. The side-face clamp of the warm-up stage was dropped after 8k; the final policy (49{,}999) uses the plain object-pose terms.

\textbf{Bucket.} The common recipe alone lifted the bucket but dropped it during the carry (28k). Adding the demonstration-gated hand--object proximity term and continuing for 12k iterations gave a policy that lifts, carries and sets the bucket down (39{,}999).

\begin{figure}[!t]
\centering
\includegraphics[width=\columnwidth]{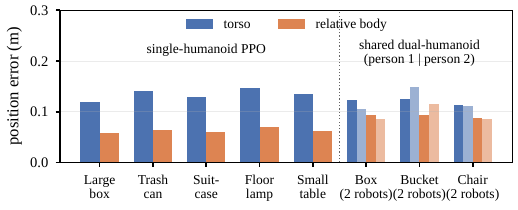}
\caption{Final training-time torso and relative-body position errors of the single-humanoid policies and of both robots of each shared dual-humanoid policy (person~1 solid, person~2 light).}
\label{fig:errors}
\end{figure}

\textbf{Statistics.} Fig.~\ref{fig:errors} compares the final errors. The two robots of each scene reach similar torso errors (box 0.124/0.106\,m, bucket 0.125/0.148\,m, chair 0.114/0.112\,m for persons 1/2), consistent with one policy serving both roles, and these lie in the range of the single-humanoid policies despite the second robot, the shared object and the additional termination rules; relative body errors are higher (0.085--0.116\,m vs.\ 0.059--0.069\,m), reflecting the arm postures imposed by the grasp. Mean episode lengths are 338, 313 and 386 steps. All three final checkpoints complete the demonstrated carry in GUI evaluation (Figs.~\ref{fig:teaser}, \ref{fig:dual} and \ref{fig:chair}).

\subsection{Case Study: Why the Chair Needed More Than the Box Recipe}
\label{sec:chair}
\begin{figure}[!t]
\centering
\includegraphics[width=0.94\columnwidth]{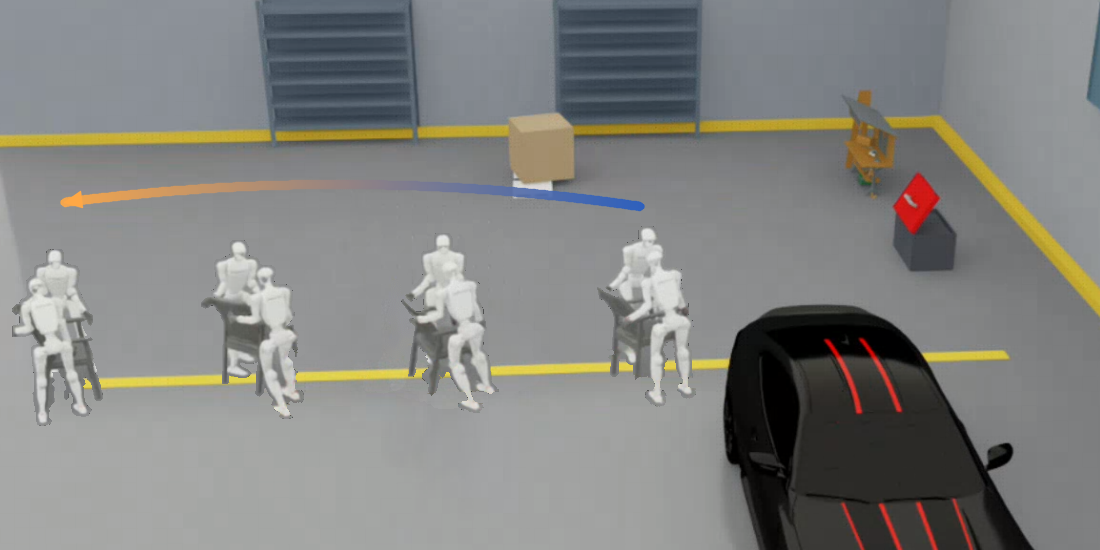}
\caption{Final shared-policy chair carry (CORE4D-Real \texttt{20231003\_2/011}, \texttt{chair006}). Four key frames of the stage-4 policy: the chair is picked up, carried and set down clear of both robots' feet. Arrow: object trajectory from start (blue) to end (orange).}
\label{fig:chair}
\end{figure}
The chair is the only class whose recipe differs from the common one in more than one additive term, and every change was driven by a diagnosed failure.

\emph{Stage~1 (0--20k), bucket recipe.} Trained from scratch with the hand-proximity term and 0.45\,m gates, the policy lifted the chair and let it slip (16k), then carried it for a while and dropped it (20k). The references explain why (Fig.~\ref{fig:refs}, bottom): the demonstrated wrists stay 0.46--0.50\,m from the chair centre, so the soft gate was at most 0.38 and the contacting-wrist termination never armed---neither term acted during the transport.

\emph{Stage~2 (20k--30k), gates 0.60\,m.} Widening both gates produced the first rollouts that reached the put-down (30k), but it was unstable: of 256 diagnostic scenes 8 completed, 76 terminated within the first second (spawn), 85 between 4 and 6\,s by the object-position rule and 10 after 9\,s by the torso rule. The chair was carried 0.12--0.14\,m below the demonstration, touched the ground while the reference was still moving it, and was dragged into place.

\emph{Stage~3 (30k--40k), object-height term.} A tight height term $\exp(-\Delta z^2/0.08^2)$ (weight 1.0) did not raise the carry: 7 of 256 scenes completed, the carry-phase height error stayed at $-0.12$ to $-0.13$\,m and the 4--6\,s object terminations rose from 85 to 102. At a 0.12\,m error the term is already $\approx0.1$ and has almost no gradient; more importantly, the offset is inherited from the retargeted reference, whose feet hover 0.081 and 0.087\,m above the floor while the robot stands on it, so the torso sits 0.07--0.13\,m below its reference and hands and chair follow (in the object frame the wrists match the demonstrated grip points). GUI rollouts also exposed a second failure: carries ended with the chair set down on a robot's foot. During the carry an ankle was inside the footprint expanded by 0.10\,m in 17--67\,\% (person~1) and 20--57\,\% (person~2) of surviving scenes, 5 of the 7 completed scenes ended with a foot under the chair, and the robots stood 0.05--0.14\,m closer to the chair than the demonstrated persons at put-down, who place a foot inside the expanded footprint in a single frame.

\emph{Discarded branch.} Lowering both references by 0.05\,m so that the reference feet rest on the floor, with the height term and an \emph{ungated} footprint penalty (margin 0.12\,m), halved the spawn terminations (44 vs.\ 88) and removed the feet-under-chair events, but person~2 now lagged 0.10--0.17\,m behind its reference during the approach (reference-position terminations 66\,$\to$\,108) and the chair was still carried 0.13--0.15\,m low; the branch was stopped at 45k with flat metrics (dashed in Fig.~\ref{fig:curves}).

\emph{Stage~4 (40k--50k), gated feet penalty only.} Resuming from stage~3 with the original reference, the 0.60\,m gates, the gated feet--footprint penalty of Section~\ref{sec:reward} and no height term gave the final policy (49{,}997). Its episode length (386 steps) is the longest and its torso errors the smallest of the three classes, and the mean footprint penalty is close to zero ($-0.006$ per step). In evaluation the chair is picked up, carried and set down clear of both robots' feet (Fig.~\ref{fig:chair}); it is still carried lower than in the demonstration because the retargeted foot height, the actual source of the offset, was not corrected.

\textbf{Observations.} Termination mattered as much as reward: the torso-XY and contacting-wrist rules removed the degenerate strategies that height-only checks permit in a two-robot scene (Fig.~\ref{fig:curves}), and every later term was demonstration-gated so that it could not interfere with the approach. The common object-pose reward is not specific enough when the grasp is far from the object centre: with $\sigma=0.3$\,m it still pays about 85\,\% of its maximum for a chair carried 0.12\,m too low, and the bucket was dropped until the horizontal wrist slack was rewarded directly. Finally, failures were diagnosed from kinematic quantities the pipeline already computes---wrist, pelvis and foot positions in the object frame and the termination reason and time---not from reward curves, which were flat or improving in every discarded branch.

\section{Limitations and Conclusion}
\enlargethispage{3\baselineskip}
The evaluation is qualitative: success is established by repeated GUI rollouts of one scene, and the quantitative numbers are training-time statistics and stage-comparison diagnostics under the training randomization rather than success rates over many clean episodes. Each policy is trained on one clip of one object and is not expected to generalize to other clips, objects or partners; the shared policy is specialized to two identical robots; all results are in simulation with a 0.1--0.4\,kg object and without sensor noise. The chair result inherits a retargeting artifact (hovering reference feet) that training compensated for but did not remove; a per-frame ground-contact correction in retargeting would address the cause.

We described an end-to-end pipeline that converts human--object and human--object--human motion capture into whole-body tracking policies for the 29-DoF Unitree G1, validated the single-humanoid path on eight OMOMO objects, and extended it to two-humanoid carrying with one shared 29-DoF policy sampled twice per scene and coupled only through the simulated object. A box, a bucket and a chair from CORE4D were learned with a common tracking reward plus a few demonstration-gated, class-conditional terms. A new clip of an existing class needs only its sequence identifier, mesh and reset height; a new class needs a choice of which documented terms its contact geometry requires. Code, presets and stage commands: \url{https://github.com/Ruhsjw/omomo-dual-g1-carry}.

\balance
\bibliographystyle{IEEEtran}
\bibliography{references}

\end{document}